\documentclass[a4paper,english]{ecai2008}
\usepackage[T1]{fontenc}
\usepackage[latin9]{inputenc}
\usepackage{float}
\usepackage{amsthm}
\usepackage{graphicx}
\usepackage{setspace}
\usepackage{amssymb}

\makeatletter

\makeatletter

\usepackage{times}
\usepackage{latexsym}
\usepackage{babel}
\usepackage{algorithm}
\usepackage[noend]{algpseudocode}

\begin{document}

\newcommand{\noun}[1]{\textsc{#1}}

\makeatother

\title{An Efficient Diagnosis Algorithm for\\ Inconsistent Constraint Sets\footnote{Preprint of: A. Felfernig, M. Schubert, and C. Zehentner. An Efficient Diagnosis Algorithm for Inconsistent Constraint Sets. Artificial Intelligence for Engineering Design, Analysis, and Manufacturing (AIEDAM), Cambridge University Press, vol. 26, no.1, pp. 53-62, 2012.}}
  \author{Alexander Felfernig$^2$ \and Monika Schubert$^2$ \and Christoph Zehentner
\thanks{TU Graz, Institute of Software Technology, Applied Software Engineering \& AI, Austria, email: \{felfernig, schubert, zehentner\}@ist.tugraz.at
}}
\maketitle

\begin{abstract}
Constraint sets can become inconsistent in different contexts. For
example, during a configuration session the set of customer requirements
can become inconsistent with the configuration knowledge base. Another
example is the engineering phase of a configuration knowledge base
where the underlying constraints can become inconsistent with a set
of test cases. In such situations we are in the need of techniques
that support the identification of minimal sets of faulty constraints
that have to be deleted in order to restore consistency. In this paper
we introduce a divide-and-conquer based diagnosis algorithm (\noun{FastDiag})
which identifies minimal sets of faulty constraints in an over-constrained
problem. This algorithm is specifically applicable in scenarios where
the efficient identification of leading (preferred) diagnoses is crucial.
We compare the performance of \noun{FastDiag} with the conflict-directed
calculation of hitting sets and present an in-depth performance analysis
that shows the advantages of our approach.
\end{abstract}


\maketitle

\section{Introduction\label{sec:INTRODUCTION}}

Constraint technologies \cite{Tsang1993} are applied in different
areas such as configuration \cite{Mittal1989,FleischanderlIntelligentSystems1998,Sinz2007},
recommendation \cite{FelfernigIJCAI2009}, and scheduling \cite{Castillo2005}.
There are many scenarios where the underlying constraint sets can
become over-constrained. For example, when implementing a configuration
knowledge base, constraints can become inconsistent with a set of
test cases \cite{FelfernigAIJournal2004}. Alternatively, when interacting
with a configurator application \cite{OSullivanAAAI2007,FelfernigIJCAI2009},
the given set of customer requirements (represented as constraints)
can become inconsistent with the configuration knowledge base. In
both situations there is a need of an intelligent assistance that
actively supports users of a constraint-based application (end users
or knowledge engineers). A wide-spread approach to support users in
the identification of minimal sets of faulty constraints is to combine
conflict detection \noun{(}see, e.g., \cite{JunkerAAAI2004}) with
a corresponding hitting set algorithm \cite{deKleerWilliams1987,ReiterAIJournal1987,deKleerAIJournal1992}.
In their original form these algorithms are applied for the calculation
of \emph{minimal }(\emph{cardinality})\emph{ diagnoses} which are
typically determined with breadth-first search. Further diagnosis
algorithms have been developed that follow a best-first search regime
where the expansion of the hitting set search tree is guided by failure
probabilities of components \cite{deKleerAIJournal1990}. Another
example for such an approach is presented in \cite{FelfernigIJCAI2009}
where similarity metrics are used to guide the (best-first) search
for a preferred (plausible) minimal diagnosis (including repairs). 

Both, simple breadth-first search and best-first search diagnosis
approaches are predominantly relying on the calculation of conflict
sets \cite{JunkerAAAI2004}. In this context, the determination of
a minimal diagnosis of cardinality \emph{n} requires the identification
of at least \emph{n} minimal conflict sets. In this paper, we introduce
a direct diagnosis algorithm (\noun{FastDiag}) that allows to determine \emph{one
minimal diagnosis at a time} with the same computational effort needed for determining \emph{one conflict set at a time}. \noun{FastDiag}
supports the identification of preferred  diagnoses given predefined
preferences regarding a set of decision alternatives. It
boosts the applicability of diagnosis methods in scenarios such
as online configuration \& reconfiguration \cite{FelfernigAIJournal2004},
recommendation of products \& services \cite{FelfernigIJCAI2009},
and (more generally) in scenarios where the efficient calculation
of preferred (leading) diagnoses is crucial \cite{deKleerAIJournal1990}.
\noun{FastDiag} is not restricted to constraint-based systems but
it is also applicable, for example, in the context of SAT solving
\cite{MarquesSilva1996} and description logics reasoning \cite{FriedrichISWC2005}.

The remainder of this paper is organized as follows. In Section \ref{sec:WORKING-EXAMPLE}
we introduce a simple example configuration task  from the automotive
domain. In Section \ref{sec:DIAGNOSING-FAULTY-CONSTRAINT} we discuss
the basic hitting set based approach to the calculation of diagnoses.
In Section \ref{sec:CALCULATING-PLAUSIBLE-DIAGNOSES} we introduce
an algorithm (\noun{FastDiag}) for calculating preferred\emph{ }diagnoses
for a given over-constrained problem. In Section \ref{sec:EVALUATION}
we present a detailed evaluation of \noun{FastDiag} which clearly
outperforms standard hitting set based algorithms in the calculation
of the \emph{topmost}-\emph{n} preferred diagnoses. With Section \ref{sec:RELATED-WORK}
we provide an overview of related work in the field. The paper is
concluded with Section \ref{sec:CONCLUSION}.

\section{Example Domain: Car Configuration\label{sec:WORKING-EXAMPLE}}

Car configuration will serve as a working example throughout this
paper. Since we exploit configuration problems  for the discussion
of our diagnosis algorithm, we first introduce a formal definition
of a configuration task. This definition is based on \cite{FelfernigAIJournal2004}
but is given in the context of a constraint satisfaction problem (CSP)
\cite{Tsang1993}.

\textbf{Definition 1 (Configuration Task)}. A configuration task can
be defined as a CSP (V, D, C). V = \{v$_{1}$, v$_{2}$, \ldots{},
v$_{n}$\} represents a set of finite domain variables. D = \{dom(v$_{1}$),
dom(v$_{2}$), \ldots{}, dom(v$_{n}$)\} represents a set of variable
domains dom(v$_{k}$) where dom(v$_{k}$) represents the domain of
variable v$_{k}$. C = C$_{KB}$ $\cup$ C$_{R}$ where C$_{KB}$
= \{c$_{1}$, c$_{2}$, \ldots{}, c$_{q}$\} is a set of domain specific
 constraints (the configuration knowledge base) that restrict the
possible combinations of values assigned to the variables in V. C$_{R}$
= \{c$_{q+1}$, c$_{q+2}$, \ldots{}, c$_{t}$\} is a set of customer
requirements also represented as constraints.

A simplified example of a configuration task in the automotive domain
is the following. In this example, \emph{type} represents the car
type, \emph{pdc} is the parc distance control functionality, \emph{fuel}
represents the fuel consumption per 100 kilometers, a \emph{skibag}
allows ski stowage inside the car, and \emph{4-wheel} represents the
corresponding actuation type. These variables describe the potential
set of requirements that can be specified by the user (customer).
The possible combinations of these requirements are defined by a set
of constraints which are denoted as \emph{configuration knowledge
base} (C$_{KB}$) which is defined as C$_{KB}$ = \{c$_{1}$, c$_{2}$,
c$_{3}$, c$_{4}$\} in our example. Furthermore, we assume the set
of \emph{customer requirements} C$_{R}$ = \{c$_{5}$, c$_{6}$, c$_{7}$\}.%
\footnote{Note that constraints are not necessarily \emph{unary} or \emph{binary}
(we tried to keep the example simple), they can also be \emph{n-ary}.%
}
\begin{itemize}
\item $V$ = \{\textbf{\emph{type}},\emph{ }\textbf{\emph{pdc}},\emph{ }\textbf{\emph{fuel}},\textbf{\emph{
skibag}},\emph{ }\textbf{\emph{4-wheel}}\}
\item $D$ = \{dom(\textbf{\emph{type}})=\{\emph{city}, \emph{limo}, \emph{combi},
\emph{xdrive}\}, dom(\textbf{\emph{pdc}})= \{\emph{yes}, \emph{no}\},
dom(\textbf{\emph{fuel}}) = \{\emph{4l},\emph{ 6l},\emph{ 10l}\},
dom(\textbf{\emph{skibag}})=\{\emph{yes}, \emph{no}\}, dom(\textbf{\emph{4-wheel}})=\{\emph{yes},
\emph{no}\}
\item $C_{KB}$ = \{$c_1$: \textbf{\emph{4-wheel}} = \emph{yes} $\Rightarrow$
\textbf{\emph{type}} = \emph{xdrive}, $c_2$: \textbf{\emph{skibag}}
= \emph{yes} $\Rightarrow$ \textbf{\emph{type}} $\neq$ \emph{city},
$c_3$: \textbf{\emph{fuel}} = \emph{4l} $\Rightarrow$ \textbf{\emph{type}}
= \emph{city}, $c_4$: \textbf{\emph{fuel}} = \emph{6l} $\Rightarrow$
\textbf{\emph{type}} $\neq$ \emph{xdrive}\}
\item $C_R$ = \{$c_5$: \textbf{\emph{type}} = \emph{combi}, $c_6$: \textbf{\emph{fuel}}
= \emph{4l}, $c_7$: \textbf{\emph{4-wheel}} = yes\}
\end{itemize}
On the basis of this configuration task definition, we can now introduce
the definition of a concrete \emph{configuration} (solution for a
configuration task).

\textbf{Definition 2 (Configuration)}. A configuration for a given
configuration task (V, D, C) is an instantiation I = \{v$_{1}$=ins$_{1}$,
v$_{2}$=ins$_{2}$, \ldots{}, v$_{n}$=ins$_{n}$\} where ins$_{k}$
$\in$ dom(v$_{k}$).

A configuration is \emph{consistent} if the assignments in I are consistent
with the c$_{i}$ $\in$ C. Furthermore, a configuration is \emph{complete}
if all variables in V are instantiated. Finally, a configuration is
\emph{valid} if it is consistent and complete.

\section{Diagnosing Over-Constrained Problems\label{sec:DIAGNOSING-FAULTY-CONSTRAINT}}

For the configuration task  introduced in Section \ref{sec:WORKING-EXAMPLE}
we are \emph{not} able to find a solution, for example, a \emph{combi}-type
car does not support a fuel consumption of \emph{4l per 100 kilometers}.
Consequently, we want to identify minimal sets of constraints (c$_{i}$
$\in$ C$_{R}$) which have to be deleted  in order to be able to
identify a solution (restore the consistency). In the example of Section
\ref{sec:WORKING-EXAMPLE} the set of constraints C$_{R}$=\{c$_{5}$,
c$_{6}$, c$_{7}$\} is inconsistent with the constraints C$_{KB}$=
\{c$_{1}$, c$_{2}$, c$_{3}$, c$_{4}$\}, i.e., no solution can
be found for the underlying configuration task. A standard approach
to determine a minimal set of constraints that have to be deleted
from an over-constrained problem is to resolve all minimal conflicts
contained in the constraint set. The determination of such constraints
is based on a conflict detection algorithm (see, e.g., \cite{JunkerAAAI2004}),
the derivation of the corresponding diagnoses is based on the calculation
of hitting sets \cite{ReiterAIJournal1987}. Since both, the notion
of a (\emph{minimal}) \emph{conflict} and the notion of a (\emph{minimal})
\emph{diagnosis} will be used in the following sections, we provide
the corresponding definitions here. 

\textbf{Definition 3 (Conflict Set)}. A conflict set is a set CS $\subseteq$
C$_{R}$ s.t. C$_{KB}$ $\cup$ CS is inconsistent. CS is a \emph{minimal}
if there does not exist a conflict set CS\textquoteright{} with CS\textquoteright{}
$\subset$ CS. 

In our working example we can identify three minimal conflict sets
which are CS$_{1}$=\{c$_{5}$,c$_{6}$\}, CS$_{2}$=\{c$_{5}$,c$_{7}$\},
and CS$_{3}$=\{c$_{6}$,c$_{7}$\}. 

CS$_{1}$, CS$_{2}$, CS$_{3}$ are conflict sets since CS$_{1}$
$\cup$ C$_{KB}$ $\vee$ CS$_{2}$ $\cup$ C$_{KB}$ $\vee$ CS$_{3}$
$\cup$ C$_{KB}$ is inconsistent. The minimality property is fulfilled
since there does not exist a conflict set CS$_{4}$ with CS$_{4}$
$\subset$ CS$_{1}$ or CS$_{4}$ $\subset$ CS$_{2}$ or CS$_{4}$
$\subset$ CS$_{3}$. The standard approach to resolve the given conflicts
is the construction of a corresponding \emph{hitting set} \emph{directed
acyclic graph} (\emph{HSDAG}) \cite{ReiterAIJournal1987} where the
resolution of all minimal conflict sets automatically corresponds
to the identification of a minimal diagnosis. A minimal diagnosis
in our application context is a minimal set of customer requirements
contained in the set of car features (C$_{R}$) that has to be deleted
from C$_{R}$  in order to make the remaining constraints consistent
with C$_{KB}$. Since we are dealing with the diagnosis of customer
requirements, we introduce the definition of a \emph{customer requirements
diagnosis problem} (Definition 4). This definition is based on the
definition given in \cite{FelfernigAIJournal2004}.

\textbf{Definition 4 (CR Diagnosis Problem)}. A customer requirements
diagnosis (CR diagnosis) problem is defined as a tuple (C$_{KB}$,
C$_{R}$) where C$_{R}$ is the set of given customer requirements
and C$_{KB}$ represents the constraints part of the configuration
knowledge base.

The definition of a \emph{CR diagnosis} that corresponds to a given
CR Diagnosis Problem is the following (see Definition 5).

\textbf{Definition 5 (CR Diagnosis)}. A CR diagnosis for a CR diagnosis
problem (C$_{KB}$, C$_{R}$) is a set $\Delta$ $\subseteq$ C$_{R}$,
s.t., C$_{KB}$ $\cup$ (C$_{R}$ - $\Delta$) is consistent. $\Delta$
is \emph{minimal} if there does not exist a diagnosis $\Delta$\textquoteright{}
$\subset$ $\Delta$ s.t. C$_{KB}$ $\cup$ (C$_{R}$ - $\Delta$\textquoteright{})
is consistent.

The HSDAG algorithm for determining minimal diagnoses is discussed
in detail in \cite{ReiterAIJournal1987}. The concept of this algorithm
will be explained on the basis of our working example. It relies on
a conflict detection algorithm that is responsible for detecting minimal
conflicts in a given set of constraints (in our case in the given
customer requirements). One conflict detection algorithm is \noun{QuickXplain}
\cite{JunkerAAAI2004} which is based on an efficient divide-and-conquer
search strategy. For the purposes of our working example let us assume
that the first minimal conflict set determined by \noun{QuickXplain}
is the set CS$_{1}$= \{c$_{5}$, c$_{6}$\}. Due to the minimality
property, we are able to resolve each conflict by simply deleting
one element from the set, for example, in the case of CS$_{1}$ we
have to either delete c$_{5}$ or c$_{6}$. Each variant to resolve
a conflict set is represented by a specific path in the corresponding
HSDAG -- the HSDAG for our working example is depicted in Figure 1.
The deletion of c$_{5}$ from CS$_{1}$ triggers the calculation of
another conflict set CS$_{3}$ = \{c$_{6}$, c$_{7}$\} since C$_{R}$
- \{c$_{5}$\} $\cup$ C$_{KB}$ is inconsistent. If we decide to
delete c$_{6}$ from CS$_{1}$, C$_{R}$ - \{c$_{6}$\} $\cup$ C$_{KB}$
remains inconsistent which means that \noun{QuickXplain} returns another
minimal conflict set which is CS$_{2}$ = \{c$_{5}$, c$_{7}$\}. 

The original HSDAG algorithm \cite{ReiterAIJournal1987} follows
a strict breadth-first search regime. Following this strategy, the
next node to be expanded in our working example is the minimal conflict
set CS$_{3}$ which has been returned by \noun{QuickXplain} for C$_{R}$
- \{c$_{5}$\} $\cup$ C$_{KB}$. In this context, the first option
to resolve CS$_{3}$ is to delete c$_{6}$. This option is a valid
one and $\Delta$$_{1}$= \{c$_{5}$, c$_{6}$\} is the resulting
minimal diagnosis. The second option for resolving CS$_{3}$ is to
delete the constraint c$_{7}$. In this case, we have identified the
next minimal diagnosis $\Delta$$_{2}$ = \{c$_{5}$, c$_{7}$\} since
C$_{R}$ - \{c$_{5}$, c$_{7}$\} $\cup$ C$_{KB}$ is consistent.
This way we are able to identify all minimal sets of constraints $\Delta$$_{i}$
that -- if deleted from C$_{R}$ -- help to restore the consistency
with C$_{KB}$. If we want to calculate the complete set of diagnoses
for our working example, we still have to resolve the conflict set
CS$_{2}$. The first option to resolve CS$_{2}$ is to delete c$_{5}$
-- since \{c$_{5}$, c$_{6}$\} has already been identified as a minimal
diagnosis, we can close this node in the HSDAG. The second option
to resolve CS$_{2}$ is to delete c$_{7}$. In this case we have determined
the third minimal diagnosis which is $\Delta$$_{3}$ = \{c$_{6}$,
c$_{7}$\}.

In our working example we are able to enumerate all possible diagnoses
that help to restore consistency. However, the calculation of all
minimal diagnoses is expensive and thus in many cases not practicable
for interactive settings. Since users are often interested in a reduced
subset of all the potential diagnoses, alternative algorithms are
needed that are capable of identifying preferred diagnoses \cite{ReiterAIJournal1987,deKleerAIJournal1990,FelfernigIJCAI2009}.
Such approaches have already been developed \cite{deKleerAIJournal1990,FelfernigIJCAI2009},
however, they are still based on the resolution of conflict sets which
is computationally expensive (see Section \ref{sec:EVALUATION}).
Our idea presented in this paper is a diagnosis algorithm that helps
to determine preferred  diagnoses without the need of calculating
 conflict sets. The basic properties of \noun{FastDiag}
will be discussed in Section \ref{sec:CALCULATING-PLAUSIBLE-DIAGNOSES}.


%
\begin{figure*}[ht]
\begin{singlespace}
\begin{centering}
\includegraphics[scale=0.5]{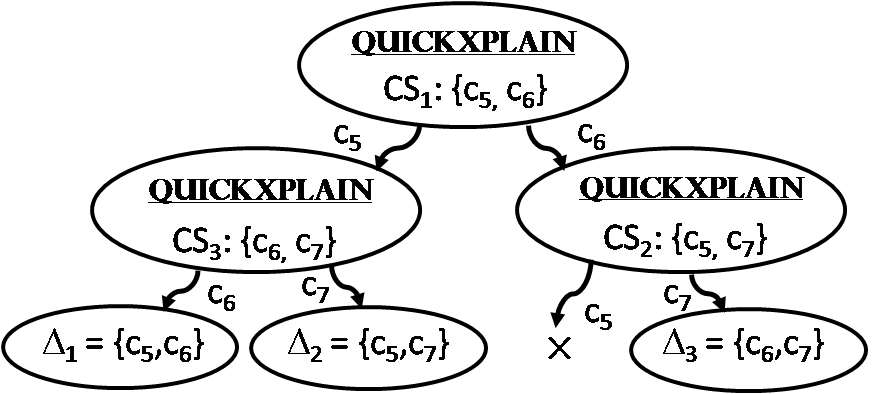}
\par\end{centering}
\end{singlespace}

\caption{HSDAG (Hitting Set Directed Acyclic Graph) \cite{ReiterAIJournal1987}
for the CR diagnosis problem (C$_{R}$=\{c$_{5}$, c$_{6}$, c$_{7}$\},
C$_{KB}$=\{c$_{1}$, c$_{2}$, c$_{3}$, c$_{4}$\}). The sets \{c$_{5}$,
c$_{6}$\}, \{c$_{6}$, c$_{7}$\}, and \{c$_{5}$, c$_{7}$\} are
the minimal diagnoses -- the conflict sets CS$_{1}$, CS$_{2}$, and
CS$_{3}$ are determined on the basis of \noun{QuickXplain} \cite{JunkerAAAI2004}.}

\end{figure*}

\section{Calculating Preferred Diagnoses with \noun{FastDiag}\label{sec:CALCULATING-PLAUSIBLE-DIAGNOSES}}

\paragraph{Preferred Diagnoses.}

Users typically prefer to keep the important requirements and to change
or delete (if needed) the less important ones \cite{JunkerAAAI2004}.
The major goal of (model-based) diagnosis tasks is to identify the
preferred (leading) diagnoses which are not necessarily minimal cardinality
ones \cite{deKleerAIJournal1990}. For the characterization of a
preferred diagnosis we will rely on the definition of a total ordering
of the given set of constraints in C (respectively C$_{R}$). Such
a total ordering can be achieved, for example, by \emph{directly asking}
the customer regarding the preferences, by applying \emph{multi-attribute
utility theory} \cite{Winterfeld1986,Ardissono2003} where the determined
interest dimensions correspond with the attributes of C$_{R}$ or
by applying the rankings determined by \emph{conjoint analysis} \cite{Belanger2005}.
The following definition of a \emph{lexicographical ordering} (Definition
6) is based on total orderings for constraints that has been applied
in \cite{JunkerAAAI2004} for the determination of \emph{preferred
conflict sets}. 

\textbf{Definition 6 (Total Lexicographical Ordering)}. Given a total
order < on C, we enumerate the constraints in C in increasing < order
c$_{1}$.. c$_{n}$ starting with the \emph{least important constraints}
(i.e., c$_{i}$ < c$_{j}$ $\Rightarrow$ i < j). We compare two subsets
X and Y of C lexicographically: 

\begin{center}
X >$_{lex}$ Y iff
\par\end{center}

\begin{center}
$\exists$k: c$_{k}$ $\in$ Y - X and
\par\end{center}

\begin{center}
X $\cap$ \{c$_{k+1}$, ..., c$_{t}$\} = Y $\cap$ \{c$_{k+1}$,
..., c$_{t}$\}.
\par\end{center}

Based on this definition of a lexicographical ordering, we can now
introduce the definition of a \emph{preferred diagnosis}.

\textbf{Definition 7 (Preferred Diagnosis)}. A minimal diagnosis $\Delta$
for a given CR diagnosis problem (C$_{R}$, C$_{KB}$) is a preferred
diagnosis for (C$_{R}$, C$_{KB}$) iff there does not exist another
minimal diagnosis $\Delta'$ with $\Delta'$ >$_{lex}$ $\Delta$.

In our working example we assumed the lexicographical ordering (c$_{5}$
< c$_{6}$ < c$_{7}$), i.e., the most important customer requirement
is c$_{7}$ (the 4-wheel functionality). If we assume that $X=\{c_5, c_7\}$
and $Y=\{c_6, c_7\}$ then $Y$-$X$ = $\{c_6\}$ and $X \cap \{c_7\}$
= $Y \cap \{c_7\}$. Intuitively, $\{c_5, c_7\}$ is a preferred diagnosis
compared to $\{c_6, c_7\}$ since both diagnoses include $c_7$ but
$c_5$ is less important than $c_6$. If we change the ordering to
(c$_{7}$ < c$_{6}$ < c$_{5}$), \noun{FastDiag} would then determine
\{c$_{6}$, c$_{7}$\} as the preferred minimal diagnosis.

\paragraph{\noun{FastDiag} Approach.}

For the following discussions we introduce the set AC which is initially
set to C$_{KB}$ $\cup$ C$_{R}$ (the union of customer requirements
(C$_{R}$) and the configuration knowledge base (C$_{KB}$)) and subsequently
changed when the algorithm runs. The basic idea of the \noun{FastDiag}
algorithm (Algorithm 1) is the following.%
\footnote{In Algorithm 1 we use the set C instead of C$_{R}$ since the application
of the algorithm is not restricted to inconsistent sets of customer
requirements.%
} In our example, the set of customer requirements C$_{R}$ = \{c$_{5}$,
c$_{6}$, c$_{7}$\} includes at least one minimal diagnosis since
C$_{KB}$ is consistent and C$_{KB}$ $\cup$ C$_{R}$ is inconsistent.
In the extreme case C$_{R}$ itself represents the minimal diagnosis
which then means that \emph{all} constraints in C$_{R}$ are part
of the diagnosis, i.e., each c$_{i}$ $\in$ C$_{R}$ represents a
singleton conflict. In our case C$_{R}$ obviously does not represent
a minimal diagnosis -- the set of diagnoses in our working example
is \{$\Delta$$_{1}$ = \{c$_{5}$, c$_{6}$\}, $\Delta$$_{2}$ =
\{c$_{5}$, c$_{7}$\}, $\Delta$$_{3}$ = \{c$_{6}$, c$_{7}$\}\}
(see Section \ref{sec:DIAGNOSING-FAULTY-CONSTRAINT}). The next step
in Algorithm 1  is to divide the set of customer requirements C$_{R}$
= \{c$_{5}$, c$_{6}$, c$_{7}$\} into the two sets C$_{1}$ = \{c$_{5}$\}
and C$_{2}$ = \{c$_{6}$, c$_{7}$\}  and to check whether AC - C$_{1}$
is already consistent. If this is the case, we can omit the set C$_{2}$
since at least one minimal diagnosis can already be identified in
C$_{1}$. In our case, AC - \{c$_{5}$\} is inconsistent, which means
that we have to consider further elements from C$_{2}$. Therefore,
C$_{2}$ = \{c$_{6}$, c$_{7}$\} is divided into the sets \{c$_{6}$\}
and \{c$_{7}$\}. In the next step we can check whether AC \textendash{}
(C$_{1}$ $\cup$ \{c$_{6}$\}) is consistent \textendash{} this is
the case which means that we do not have to further take into account
\{c$_{7}$\} for determining the diagnosis. Since \{c$_{5}$\} does
not include a diagnosis but \{c$_{5}$\} $\cup$ \{c$_{6}$\} includes
a diagnosis, we can deduce that \{c$_{6}$\} must be part of the diagnosis.
The final step is to check whether AC \textendash{} \{c$_{6}$\} leads
to a diagnosis without including \{c$_{5}$\}. We  see that AC \textendash{}
\{c$_{6}$\} is inconsistent, i.e., $\Delta$ = \{c$_{5}$, c$_{6}$\}
is a minimal diagnosis for the CR diagnosis problem (C$_{R}$ = \{c$_{5}$,
c$_{6}$, c$_{7}$\}, C$_{KB}$ = \{c$_{1}$, \ldots{}, c$_{4}$\}).
An execution trace of the \noun{FastDiag} algorithm in the context
of our working example is shown in Figure 2.

\begin{algorithm} \footnotesize
\caption{$-$ \noun{FastDiag}}
\begin{algorithmic}[1]
\Procedure{FastDiag}{$C\subseteq AC, AC=\{c_1..c_t\}$} : $diagnosis~\Delta$   
    \If{$isEmpty(C)~or~inconsistent(AC-C)$} 
        \State {$return~\emptyset$}
    \Else 
        \State {$return~\noun{FD}(\emptyset,C,AC);$}
    \EndIf
\EndProcedure \vspace{0.3cm}

\Procedure{FD}{$D,C=\{c_1..c_q\},AC$} : $diagnosis~\Delta$       
    \If{$D \neq \emptyset~and~consistent(AC)$}  
        \State {$return~\emptyset$}
    \ElsIf{$singleton(C)$}  
        \State {$return~C$}
    \Else
        \State $k = \frac{q}{2}$;
        \State $C_1 = \{c_1..c_k\}; C_2 = \{c_{k+1}..c_q\};$
        \State $D_1 = \noun{FD}(C_1,C_2,AC-C_1);$
        \State $D_2 = \noun{FD}(D_1,C_1,AC-D_1);$
        \State $return(D_1 \cup D_2)$;
    \EndIf
\EndProcedure
\end{algorithmic}
\end{algorithm}



%
\begin{figure*}[ht]
\begin{singlespace}
\begin{centering}
\includegraphics[scale=0.5]{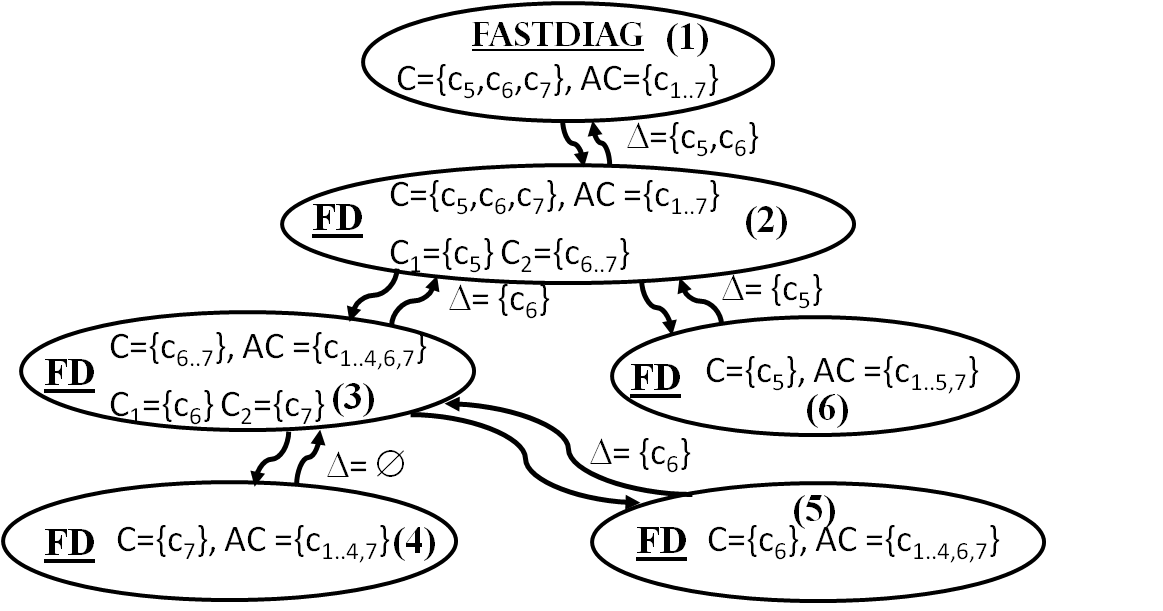}
\par\end{centering}
\end{singlespace}

\caption{\noun{FastDiag} execution trace for the CR diagnosis problem (C$_{R}$=\{c$_{5}$,
c$_{6}$, c$_{7}$\}, C$_{KB}$=\{c$_{1}$, c$_{2}$, c$_{3}$, c$_{4}$\}).
The enumerations 1--6 show the order in which the different incarnations
of the \noun{FD} function (procedure) are activated.}

\end{figure*}

\paragraph{Calculating n>1 Diagnoses.}

In order to be able to calculate \emph{n}>1 diagnoses\footnote{Typically a CR diagnosis problem has more than one related diagnosis.} with
\noun{FastDiag} we have to adopt the HSDAG construction introduced
in \cite{ReiterAIJournal1987} by substituting the resolution of
conflicts (see Figure 1) with the deletion of elements $c_i$ from
$C_R$ ($C$) (see Figure 3). In this case, a path in the HSDAG is
closed if no further diagnoses can be identified for this path or
the elements of the current path are a superset of an already closed
path (containment check). Conform to the HSDAG approach presented
in \cite{ReiterAIJournal1987}, we expand the search tree in a \emph{breadth-first}
manner. In our working example, we can delete \{c$_{5}$\} (one element
of the first diagnosis $\Delta_{1}$ = \{c$_{5}$, c$_{6}$\}) from
the set C$_{R}$ of diagnosable elements and restart the algorithm
for finding another minimal diagnosis for the CR diagnosis problem
(\{c$_{6}$, c$_{7}$\}, C$_{KB}$). Since AC - \{c$_{5}$\} is inconsistent,
we can conclude that C$_{R}$ = \{c$_{6}$, c$_{7}$\} includes another
minimal diagnosis ($\Delta_{2}$ = \{c$_{6}$, c$_{7}$\}) which is
determined by \noun{FastDiag} for the CR diagnosis problem (C$_{R}$
- \{c$_{5}$\}, C$_{KB}$). Finally, we have to check whether the
CR diagnosis problem (\{c$_{5}$, c$_{7}$\}, C$_{KB}$) leads to
another minimal diagnosis. This is the case, i.e., we have identified
the last minimal diagnosis which is $\Delta_{3}$ = \{c$_{5}$, c$_{7}$\}.
The calculation of all diagnoses in our working example on the basis
of \noun{FastDiag} is depicted in Figure 3.

Note that for a given set of constraints (C) \noun{FastDiag} always
calculates the preferred diagnosis in terms of Definition 7. If $\Delta_{1}$
is the diagnosis returned by \noun{FastDiag} and we delete one element
from $\Delta_{1}$ (e.g., c$_{5}$), then \noun{FastDiag} returns
the preferred diagnosis for the CR diagnosis problem (\{c$_{5}$,
c$_{6}$, c$_{7}$\}-\{c$_{5}$\}, \{c$_{1}$, ..., c$_{7}$\}) which
is $\Delta_{2}$ in our example case, i.e., $\Delta_{1}$>$_{lex}\Delta_{2}$.
Consequently, diagnoses part of one path in the search tree (such
as $\Delta_{1}$ and $\Delta_{2}$ in Figure 3) are in a strict preference
ordering. However, there is only a \emph{partial order }between 
diagnoses in the search tree in the sense that a diagnosis at level
\emph{k} is not necessarily preferable to a diagnosis at level \emph{k}+1. 


%
\begin{figure*}[ht]
\begin{singlespace}
\begin{centering}
\includegraphics[scale=0.5]{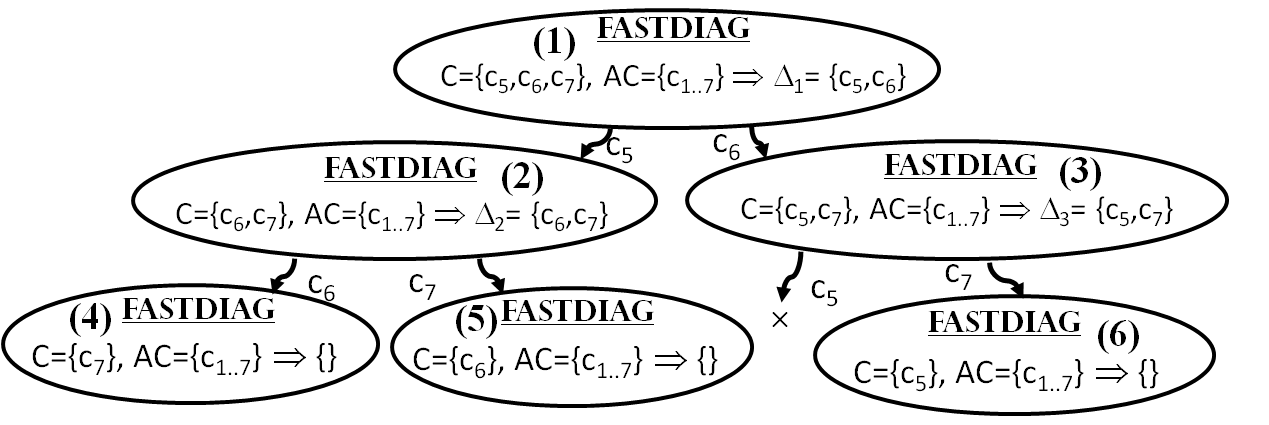}
\par\end{centering}
\end{singlespace}

\caption{\noun{FastDiag}: calculating the complete set of \emph{minimal diagnoses}.
The enumerations 1--6 show the order in which the different incarnations
of the \noun{FastDiag} algorithm are activated.}

\end{figure*}

\paragraph{\noun{FastDiag} Properties.}

A detailed listing of the basic operations of \noun{FastDiag} is shown
in Algorithm 1. First, the algorithm checks whether the constraints
in C contain a diagnosis, i.e., whether AC - C is consistent -- the
function assumes that it is activated in the case that AC is inconsistent.
If AC - C is inconsistent or C = $\varnothing$, \noun{FastDiag} returns
the empty set as result (no solution can be found -- line 2 of the
algorithm). If at least one diagnosis is contained in the set of constraints
C, \noun{FastDiag} activates the FD function (procedure) which is in charge of
retrieving a preferred  diagnosis (line 3 of the algorithm). \noun{FastDiag}
follows a divide-and-conquer strategy where the recursive function
FD divides the set of constraints (in our case the elements of C$_{R}$)
into two different subsets (C$_{1}$ and C$_{2}$) (line 8 of the
algorithm) and tries to figure out whether C$_{1}$ already contains
a diagnosis (line 5 of the algorithm). If this is the case, \noun{FastDiag}
does not further take into account the constraints in C$_{2}$. If
only one element is remaining in the current set of constraints C
and the current set of constraints in AC is still inconsistent, then
the element in C is part of a minimal diagnosis (line 6 of the algorithm).
\noun{FastDiag} is \emph{complete} in the sense that if C contains
exactly one minimal diagnosis then FD will find it. If there are multiple
minimal diagnoses then one of them (the preferred one -- see Definition
7) is returned. The recursive function FD is triggered if AC-C is
consistent and C consists of at least one constraint. In such a situation
a corresponding minimal diagnosis can be identified. If we assume
the existence of a minimal diagnosis $\Delta$ that can not be identified
by \noun{FastDiag}, this would mean that there exists at least one
constraint c$_{a}$ in C which is part of the diagnosis but not returned
by FD. The only way in which elements can be deleted from C (i.e.,
not included in a diagnosis) is by the return $\varnothing$ statement
in FD and $\varnothing$ is only returned in the case that AC is consistent
which means that the elements of C$_{2}$ (C$_{1}$) from the previous
FD incarnation are not part of the preferred diagnosis. Consequently,
it is not possible to delete elements from C which are part of the
diagnosis. \noun{FastDiag} computes only \emph{minimal diagnoses}
in the sense of Definition 5. If we assume the existence of a non-minimal
diagnosis $\Delta$ calculated by \noun{FastDiag}, this would mean
that there exists at least one constraint c$_{a}$ with $\Delta$
- \{c$_{a}$\} is still a diagnosis. The only situation in which elements
of C are added to a diagnosis $\Delta$ is if C itself contains exactly
one element. If C contains only one element (let us assume c$_{a}$)
and AC is inconsistent (in the function \noun{FD}) then c$_{a}$ is
the only element that can be deleted from AC, i.e., c$_{a}$ must
be part of the diagnosis.

\section{Evaluation\label{sec:EVALUATION}}

\paragraph{Performance of \noun{FastDiag}. }

In this section we will compare the performance of \noun{FastDiag}
with the performance of the hitting set algorithm \cite{ReiterAIJournal1987}
in combination with the \noun{QuickXplain} conflict detection algorithm
introduced in \cite{JunkerAAAI2004}. 

The worst case complexity of \noun{FastDiag} in terms of the number
of consistency checks needed for calculating one minimal diagnosis
is 2\emph{d}$\cdot$log$_{2}$($\frac{n}{d}$)+2\emph{d}, where \emph{d}
is the minimal diagnoses set size and \emph{n} is the number of constraints
(in C). The best case complexity is log$_{2}$($\frac{n}{d}$)+2\emph{d.}
In the worst case each element of the diagnosis is contained in a
different path of the search tree: log$_{2}$($\frac{n}{d}$) is the
depth of the path, 2\emph{d }represents the branching factor and the
number of leaf-node consistency checks. In the best case all elements
of the diagnosis are contained in one path of the search tree.

The worst case complexity of \noun{QuickXplain} in terms of consistency
checks needed for calculating one minimal conflict set is 2\emph{k}$\cdot$log$_{2}$($\frac{n}{k}$)+2\emph{k}
where \emph{k} is the minimal conflicts set size and \emph{n} is again
the number of constraints (in C) \cite{JunkerAAAI2004}. The best
case complexity of \noun{QuickXplain} in terms of the number of consistency
checks needed is log$_{2}$($\frac{n}{k}$)+2\emph{k} \cite{JunkerAAAI2004}.
Consequently, the number of consistency checks per conflict set (\noun{QuickXplain})
and the number of consistency checks per diagnosis (\noun{FastDiag})
fall into a logarithmic complexity class. 

Let \emph{n}$_{cs}$ be the number of minimal conflict sets in a constraint
set and \emph{n}$_{diag}$ be the number of minimal diagnoses, then
we need \emph{n}$_{diag}$ \noun{FD} calls (see Algorithm 1) plus
\emph{n}$_{cs}$ additional consistency checks and \emph{n}$_{cs}$
activations of \noun{QuickXplain} with n$_{diag}$ additional consistency
checks for determining \emph{all} diagnoses. The results of a performance
evaluation of \noun{FastDiag} are depicted in the Figures 4--7. The
basis for these evaluations was the bicycle configuration knowledge
base taken from the CLib%
\footnote{www.itu.dk/research/cla/externals/clib/%
} configuration benchmarks library (34 variables and about 65 constraints).
For this example knowledge base we randomly generated different sets
of requirements (of cardinality 5,7,10, and all possible requirements)
and measured the performance of calculating corresponding diagnosis
sets (the first diagnosis, first 5 diagnoses, first 10 diagnoses,
and all diagnoses). The runtime performance of the different diagnosis
algorithms and the needed amount of TP calls is shown in the Figures
4--7. As solver we used the CLib based decision diagram represenation
which allows for backtracking-free solution search. The tests have
been executed on a standard desktop computer (\emph{Intel(R) Core(TM)2 Quad CPU QD9400} CPU with \emph{2.66Ghz} and \emph{2GB} RAM).
Note that we have evaluated the performance of \noun{FastDiag} with
different other benchmark configuration knowledge bases on the CLib
web page with basically the same result. \noun{FastDiag} shows to
be a valuable alternative for determining diagnoses in interactive
settings especially for calculating the preferred first-n solutions.


%
\begin{figure*}[ht]
\begin{centering}
\includegraphics[scale=0.5]{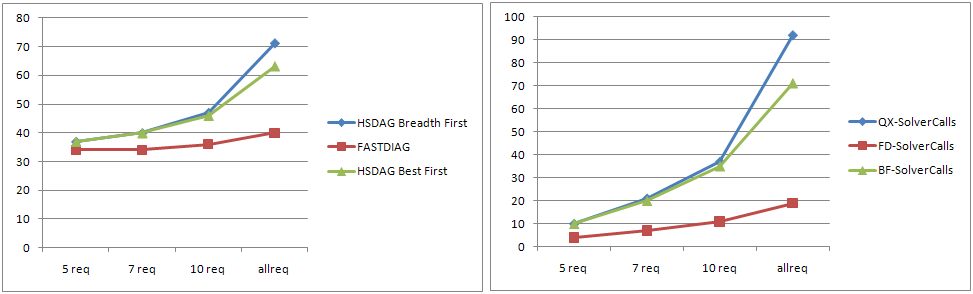}
\par\end{centering}

\caption{Calculating the \emph{first minimal diagnosis} with \noun{FastDiag}
vs. hitting set based diagnosis on the basis of \noun{QuickXplain}
for 5, 7, 10, and 15 user requirements (\emph{req}): performance in
\emph{msec} on the \emph{lhs} and \emph{number of needed TP calls}
on the \emph{rhs}.}

\end{figure*}

\begin{figure*}[ht]
\begin{centering}
\includegraphics[scale=0.5]{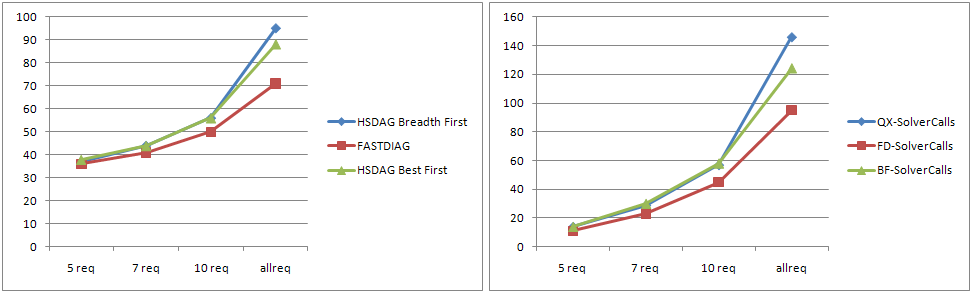}
\par\end{centering}

\caption{Calculating the \emph{topmost-5 minimal diagnoses} with \noun{FastDiag}
vs. hitting set based diagnosis on the basis of \noun{QuickXplain}
for 5, 7, 10, and 15 user requirements (\emph{req}): performance in
\emph{msec} on the \emph{lhs} and \emph{number of needed TP calls}
on the \emph{rhs}.}

\end{figure*}

\begin{figure*}[ht]
\begin{centering}
\includegraphics[scale=0.5]{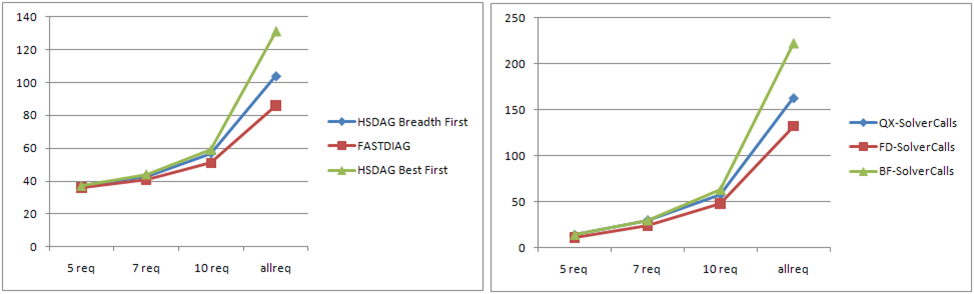}
\par\end{centering}

\caption{Calculating the \emph{topmost- 10 minimal diagnoses} with \noun{FastDiag}
vs. hitting set based diagnosis on the basis of \noun{QuickXplain}
for 5, 7, 10, and 15 user requirements (\emph{req}): performance in
\emph{msec} on the \emph{lhs} and \emph{number of needed TP calls}
on the \emph{rhs}.} 

\end{figure*}

\begin{figure*}[ht]
\begin{centering}
\includegraphics[scale=0.5]{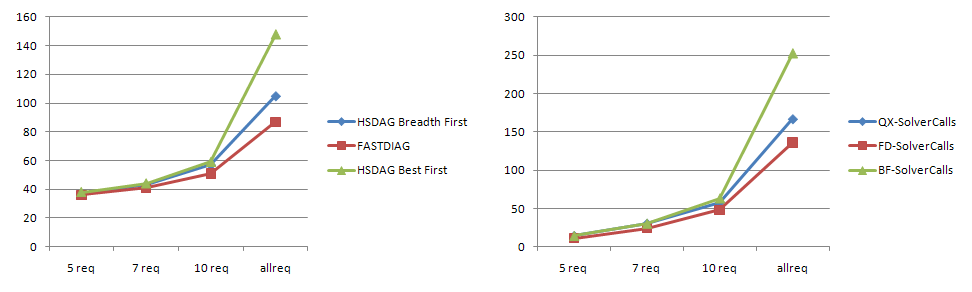}
\par\end{centering}

\caption{Calculating \emph{all minimal diagnoses} with \noun{FastDiag} vs.
hitting set based diagnosis on the basis of \noun{QuickXplain} for
5, 7, 10, and 15 user requirements (\emph{req}): performance in \emph{msec}
on the \emph{lhs} and \emph{number of needed TP calls} on the \emph{rhs}.}

\end{figure*}

Figure 4 shows a comparison between the hitting set based diagnosis
approach (denoted as \noun{HSD}AG) and the \noun{FastDiag} algorithm
(denoted as \noun{FastDiag}) in the case that only \emph{one diagnosis}
is calculated. \noun{FastDiag} clearly outperforms the \noun{HSDAG}
approach independent of the way in which diagnoses are calculated
(breadth-first or best-first). Figure 5 shows the performance evaluation
for calculating the \emph{topmost-5 minimal diagnoses}. The result
is similar to the one for calculating the first diagnosis, i.e., \noun{FastDiag}
outperforms the two \noun{HSDAG} versions. Our evaluations show that
\noun{FastDiag} is very efficient in calculating preferred minimal
diagnoses.

\paragraph{Empirical Evaluation.}

Based on a computer configuration dataset of the Graz University of
Technology (N = 415 configurations) we evaluated the three presented
approaches w.r.t. their capability of predicting diagnoses that are
acceptable for the user (diagnoses leading to selected configurations).
Each entry of the dataset consists of a set of initial \emph{user
requirements} $C_R$ inconsistent with the configuration knowledge
base $C_{KB}$ and the \emph{configuration} which had been finally
selected by the user. Since the original requirements stored in the
dataset are inconsistent with the configuration knowledge base, we
could determine those  diagnoses that indicated which minimal sets
of requirements have to be deleted in order to be able to find a solution.

We evaluated the \emph{prediction accuracy} of the three diagnosis
approaches (\emph{HSDAG breadth-first}, \noun{FastDiag}, and \emph{HSDAG
best-first}). First, we measured the distance between the \emph{predicted
position} of a diagnosis leading to a selected configuration and the
\emph{expected position of the diagnosis} (which is 1). This distance
was measured in terms of the \emph{root mean square deviation} --
RMSD (see Formula \ref{MSE}). Table \ref{tab:mse_all} depicts the
results of this first analysis. An important result is that \noun{FastDiag}
has the lowest RMSD value (0.95). Best-first HSDAG has a similar prediction
quality (RMSD = 0.97). Finally, breadth-first HSDAG has the worst
prediction quality (RMSD = 1.64).

\begin{equation}\label{MSE} RMSD = \sqrt{\frac{1}{n} \sum_{1} ^{n} (predicted~position - 1)^2} \end{equation}


%
\begin{table*}[ht]
\begin{centering}
\begin{tabular}{|c|c|c|}
\hline 
breadth-first (HSDAG) & \noun{FastDiag} & best-first (HSDAG)\tabularnewline
\hline
\hline 
1.64 & 0.95 & 0.97\tabularnewline
\hline
\end{tabular}
\par\end{centering}

\caption{Root Mean Square Deviation (RMSD) of the diagnosis approaches.\label{tab:mse_all}} 

\end{table*}

RMSD is an often used quality estimate but it provides only a limited
view on the \emph{precision} of a (diagnosis) prediction. Therefore
we wanted to analyze the precision of the diagnosis selection strategies
discussed in this paper -- a measure for the precision of a diagnosis
algorithm is depicted in Formula \ref{PRECISION}. The idea behind
this measure is to describe how often a diagnosis that leads to a
selected configuration (selected by the user) is among the topmost-n
ranked diagnoses. As shown in Table \ref{tab:Precision-of-FastDiag},
\noun{FastDiag} and \emph{best-first HSDAG} have highest prediction
accuracy in terms of precision whereas the \emph{breadth-first HSDAG}
approach shows the worst precision. 

\begin{equation} \label{PRECISION} precision = \frac{|correctly~predicted~diagnoses|}{|predicted~diagnoses|} \end{equation}

We applied a Mann-Whitney-U-Test in order to statistically analyze
differences between the three diagnosis approaches in terms of ranking
behavior. We conducted a pair wise comparison between the diagnosis
approaches on the basis of the mentioned Mann-Whitney-U-Test. We could
identify a significant difference between the rankings of \emph{best-first
HSDAG} and \emph{breadth-first HSDAG} based diagnosis ($p = 6.625e^{-5}$)
and also between \noun{FastDiag} and \emph{breadth-first HSDAG} based
diagnosis ($p < 2.441e^{-7}$). There was no significant difference
between \emph{best-first HSDAG} and \noun{FastDiag} in terms of ranking
behavior ($p = 0.12$). 


%
\begin{table*}[ht]
\begin{centering}
\begin{tabular}{|c|c|c|c|}
\hline 
top-n diagnoses & breadth-first (HSDAG) & \noun{FastDiag} & best-first (HSDAG)\tabularnewline
\hline
\hline 
n=1 & 0.51 & 0.70 & 0.74\tabularnewline
\hline 
n=2 & 0.75 & 0.88 & 0.89\tabularnewline
\hline 
n=3 & 0.87 & 0.97 & 0.96\tabularnewline
\hline
\end{tabular}
\par\end{centering}

\caption{Precision of \noun{FastDiag} vs. HSDAG based approaches.\label{tab:Precision-of-FastDiag}} 

\end{table*}

\section{Related Work\label{sec:RELATED-WORK}}

\emph{Knowledge Base Analysis}. The authors of \cite{FelfernigAIJournal2004}
introduce an algorithm for the automated debugging of configuration
knowledge bases. The idea is to combine a conflict detection algorithm
such as \noun{QuickXplain} \cite{JunkerAAAI2004} with the hitting
set algorithm used in model-based diagnosis (MBD) \cite{ReiterAIJournal1987}
for the calculation of minimal diagnoses. In this context, conflicts
are induced by test cases (examples) that, for example, stem from
previous configuration sessions, have been automatically generated,
or have been explicitly defined by domain experts. Further applications
of MBD in constraint set debugging are introduced in \cite{FelfernigAppliedIntelligence2007}
where diagnosis concepts are used to identify minimal sets of faulty
transition conditions in state charts and in \cite{FelfernigIUI2008}
where MBD is applied for the identification of faulty utility constraint
sets in the context of knowledge-based recommendation. In contrast
to \cite{FelfernigAIJournal2004,FelfernigAppliedIntelligence2007,FelfernigIUI2008},
our work provides an algorithm that allows to directly determine diagnoses
without the need to determine corresponding conflict sets. \noun{FastDiag}
can be applied in  knowledge engineering scenarios for calculating
preferred diagnoses for faulty knowledge bases given that we are able
to determine reasonable ordering for the given set of constraints
-- this could be achieved, for example, by the application of corresponding
complexity metrics \cite{Chen2003}. 

\emph{Conflict Detection}. In contrast to the algorithm presented
in this paper, calculating diagnoses for inconsistent requirements
typically relies on the existence of (minimal) conflict sets. A well-known
algorithm with a logarithmic number of consistency checks depending
on the number of constraints in the knowledge base and the cardinality
of the minimal conflicts \textendash{} \noun{QuickXplain} \cite{JunkerAAAI2004}
\textendash{} has made a major contribution to more efficient interactive
constraint-based applications. \noun{QuickXplain} is based on a divide-and-conquer
strategy. \noun{FastDiag} relies on the same principle of divide-and-conquer
but with a different focus, namely the determination of minimal diagnoses.
\noun{QuickXplain} calculates minimal conflict sets based on the assumption
of a linear preference ordering among the constraints. Similarly \textendash{}
if we assume a linear preference ordering of the constraints in C
\textendash{} \noun{FastDiag} calculates preferred diagnoses. 

\emph{Interactive Settings}. Note that in the interactive configuration
scenario discussed in this paper our goal was to support open configuration
which lets the user explore the configuration space where the system
proactively points out inconsistent requirements -- such a functionality
is often provided by commercial configuration environments. The authors
of \cite{OSullivanAAAI2007} focus on interactive settings where
users of constraint-based applications are confronted with situations
where no solution can be found. In this context, \cite{OSullivanAAAI2007}
introduce the concept of minimal exclusion sets which correspond to
the concept of minimal diagnoses as defined in \cite{ReiterAIJournal1987}.
As mentioned, the major focus of \cite{OSullivanAAAI2007} are settings
where the proposed algorithm supports users in the identification
of acceptable exclusion sets. The authors propose an algorithm (representative
explanations) that helps to improve the quality of the presented exclusion
set (in terms of diversity) and thus increases the probability of
finding an acceptable exclusion set for the user. Our diagnosis approach
calculates preferred diagnoses in terms of a predefined ordering of
the constraint set. Thus -- compared to the work of \cite{OSullivanAAAI2007}
-- we follow a different approach in terms of focusing more on preferences
than on the degree of representativeness. 

\emph{Diagnosis Algorithms}. There are a couple of algorithms that
help to improve the efficiency of diagnosis determination -- they
are further developments of the original algorithm introduced by Reiter
\cite{ReiterAIJournal1987}. These approaches focus on making the
construction of hitting sets more efficient. Wotawa \cite{Wotawa2001}
introduces an algorithm that reduces the number of subset checks compared
to the original HSDAG approach \cite{ReiterAIJournal1987}. Fijany
et al. \cite{Fijany2004} introduce an approach to represent the
problem of determining minimal hitting sets as a corresponding integer
programming problem. Further approaches to optimize the determination
of hitting sets are discussed in \cite{Lin2003}. All the mentioned
approaches rely on (minimal) conflict sets which are the basis for
calculating a set of minimal diagnoses, whereas \noun{FastDiag} is
a complete and minimal diagnosis algorithm without the need of conflict
sets. It is important to mention that especially when calculating
the first n-diagnoses (for n > 1, i.e., not a single diagnosis), \noun{FastDiag}
can also exploit the mentioned algorithms of \cite{Lin2003,Wotawa2001}
for the calculation of more than one diagnosis, i.e., it is not bound
to the usage of the original HSDAG algorithm. Lin et al. \cite{Lin2002}
introduce an approach to determine hitting sets on the basis of genetic
algorithms; a similar approach to the determination of diagnoses is
presented in \cite{Feldman2008} who introduce a stochastic fault
diagnosis algorithm which is based on greedy stochastic search. Such
approaches show to significantly improve search performance, however,
there is no general guarantee of completeness and diagnosis minimality.
Finally, there exist a couple of algorithms that are improving the
algorithmic performance of diagnosis calculation due to additional
knowledge about the structural properties of the diagnosis problem.
For example, \cite{Jannach2006} show the determination of (minimal)
diagnoses for the case of conjunctive queries on database tables (the
set of diagnoses can be precompiled by executing the individual parts
of the query on the given dataset); Siddiqi et al. \cite{Siddiqi2007}
show one approach to exploit structural properties of system descriptions
to improve the overall performance of diagnosis determination -- in
this case, \emph{cones} are areas in a gate with a certain structure
and a certain probability of including a diagnosis -- the search process
focuses on exactly those areas. \noun{FastDiag} does not exploit specific
properties of the underlying constraint set, however, taking into
account such properties can further improve the performance of the
algorithm -- corresponding evaluations are within the scope of future
work.

\emph{Personalized Diagnosis}. Many of the existing diagnosis approaches
do not take into account the need for personalizing the set of diagnoses
to be presented to a user. Identifying diagnoses of interest in an
efficient manner is a clear surplus regarding the acceptance of the
underlying application, for example, users of a configurator application
are not necessarily interested in minimal cardinality diagnoses \cite{ReiterAIJournal1987}
but rather in those that correspond to their current preferences.
A first step towards the application of personalization concepts in
the context of knowledge-based recommendation is presented in \cite{FelfernigIJCAI2009}.
The authors introduce an approach that calculates leading diagnoses
on the basis of similarity measures used for determining n-nearest
neighbors. A general approach to the identification of preferred diagnoses
is introduced in \cite{deKleerAIJournal1990} where probability estimates
are used to determine the leading diagnoses with the overall goal
to minimize the number of measurements needed for identifying a malfunctioning
device. Basic principles of determining diagnoses in knowledge-based
recommendation scenarios are discussed in \cite{Jannach2006}. Furthermore,
\cite{Froehlich94} introduce a logical characterization of preferences
which are expressed as preference relations on single diagnoses and
modal logical formulas on groups of diagnoses. In contrast to our
work, \cite{Froehlich94} do not provide an algorithm to efficiently
calculate preferred diagnoses. We see our work as a major contribution
in this context since \noun{FastDiag} helps to identify leading diagnoses
more efficiently -- further empirical studies in different application
contexts are within the major focus of our future work.

\section{Conclusion\label{sec:CONCLUSION}}

In this paper we have introduced a new diagnosis algorithm (\noun{FastDiag})
which allows the efficient calculation of one diagnosis at a time
with logarithmic complexity in terms of the number of consistency
checks. Thus, the computational complexity for the calculation of
one minimal diagnosis is equal to the calculation of one minimal conflict
set in hitting set based diagnosis approaches. The algorithm is especially
applicable in settings where the number of conflict sets is equal
to or larger than the number of diagnoses, or in settings where preferred
(leading) diagnoses are needed. Issues for future work are the determination
of repair actions for diagnoses, the further development of \noun{FastDiag}
for supporting anytime diagnosis tasks, and the conduction of further
empirical studies in different configurator application domains.

\bibliographystyle{plain}
\bibliography{references}

\end{document}